\documentclass{bioinfo}
\copyrightyear{2018} \pubyear{2018}
\usepackage{float}
\access{Advance Access Publication Date: Day Month Year}
\appnotes{xxxxxxxxx}

\begin{document}
\firstpage{1}

\subtitle{Sequence Analysis}

\title[Gender Determination with CNN]{Child Gender Determination with Convolutional Neural Networks on Hand Radio-Graphs}
\author[Mumtaz A. Kaloi \textit{et~al}.]{Mumtaz A. Kaloi\,$^{\text{\sfb 1,2}}$,  Kun He\,$^{\text{\sfb 1,}*}$}
\address{$^{\text{\sf 1}}$School of Computer Science and Technology, Huazhong University of Science and Technology, Wuhan, 430074, China and \\
$^{\text{\sf 2}}$Department of Computer Systems Engineering, Sukkur IBA University, Sukkur, 65200,
Pakistan.}
\corresp{$^\ast$To whom correspondence should be addressed.}
\history{Received on XXXXX; revised on XXXXX; accepted on XXXXX}
\editor{Associate Editor: XXXXXXX}
\abstract{\textbf{Motivation:} In forensic or medico-legal investigation as well as in anthropology the gender determination of the subject (hit by a disastrous or any kind of traumatic situation) is mostly the first step. In state-of-the-art techniques the gender is determined by examining dimensions of the bones of skull and the pelvis area. In worse situations when there is only a small portion of the \textit{human remains} to be investigated and the subject is a child, we need alternative techniques to determine the gender of the subject. In this work we propose a technique called GDCNN (Gender Determination with Convolutional Neural Networks), where the left hand radio-graphs of the children between a wide range of ages in 1 month to 18 years are examined to determine the gender. To our knowledge this technique is first of its kind. Further to identify the area of the attention we used Class Activation Mapping (CAM).\\
\textbf{Results:} The results suggest the accuracy of the model is as high as 98\%, which is very convincing by taking into account the incompletely grown skeleton of the children. The attention observed with CAM discovers that the lower part of the hand around carpals (wrist) is more important for child gender determination.\\
\textbf{Availability:} \\
\textbf{Contact:}~\href{mumtaz.ali@iba-suk.edu.pk}{mumtaz.ali@iba-suk.edu.pk}, \href{brooklet60@hust.edu.cn}{brooklet60@hust.edu.cn}, \\
\textbf{Supplementary information:} Supplementary data will be available on Github soon.}
\maketitle
\section{Introduction}

Anthropometry is the scientific approach to measure and quantify the human body. The measurements can be carried out on a living person, a skeleton or a dead body of a human being. In forensic investigation the identity of the deceased becomes a huge challenge when the subject is not only a minor but also a victim of a disastrous situation. The incidents which occur due to a criminal and deliberate act, in such situations usually the \textit{remains} of the victims are destroyed by the culprits to mislead the forensic investigators in identification of the deceased. To deal with such situations there is need of a largely automated mechanism. Most common situations for instance aircraft and car crashes, bomb blasts, chemical explosions, and other fatalities like fire eruptions may worsen the job for forensic investigators. Therefore automation of the forensic investigation systems is an ultimate and absolute goal of the modern world forensics. During the forensic investigation mostly the gender, age, race and stature are examined to identify the victim at initial stage \citep{david17}. Determination of the gender by incorporating anthropometric measurements is a research area which involves the measurements of hands, feet, and measurements of the extremities (upper, lower and long bones) in the body \citep{aboulh11}, \citep{ibrahim16}. 

The gender determination based on \textit{human remains} usually finds its foundations in sexual dimorphism and most of the bones in human body may contain its presence up to a certain degree. However this method is not always suitable mainly due to the possibility of certain significant chemical and physical damage to the \textit{remains} of the subject. In second approach the most common and popular parts of human body for gender determination are skull and the pelvis area \citep{scdpin16}. 

\subsection{Previous Work of Gender Determination}
Before this research a number of researchers have tried different techniques to determine the gender (mostly the gender of adults). For instance~\citep{ibrahim16} investigated the hand dimensions of 600 individuals. They used length, breadth and hand dimensions. They also investigated the ratio between index and ring fingers. They claimed in their results that the ratio of index and ring finger is higher in females as compared to males.~\citep{aboulh11} also determined the gender by calculating hand dimensions, index finger and ring finger ratio in upper Egyptians. Their dataset consisted of 250 males and 250 females. All the subjects were adults over 18 years old. They claimed in their results that the average male hands are 1.3 cm larger than females. They also suggested thresholds for index and ring finger lengths to determine the gender.

\citep{thomas18}~investigated how to determine the gender of individuals by Maxillary Sinus Volumes (MSV). They used one hundred and three CT scan images. They determined the MSV by 3D reconstructions. They by their results suggested that MSV establishment can be useful to determine the gender of individuals.
 \citep{fgiu13} used the CT scans of the skulls of 200 patients for estimating the scapular diameters. They carried out their estimation by analyzing scapulae of each patient after 3D post processing reconstructions. They claimed to achieve accuracy of 88\% on the calibration sample, and their system correctly identified 9 out of 10 in testing samples.
 \citep{edwards13} presented a method to investigate the values of measurements of the foramen magnum in relation to the determination of biological gender. They investigated a sample of 250 adults by obtaining CT scans.
 \begin{figure}[H]
\includegraphics[width=\linewidth]{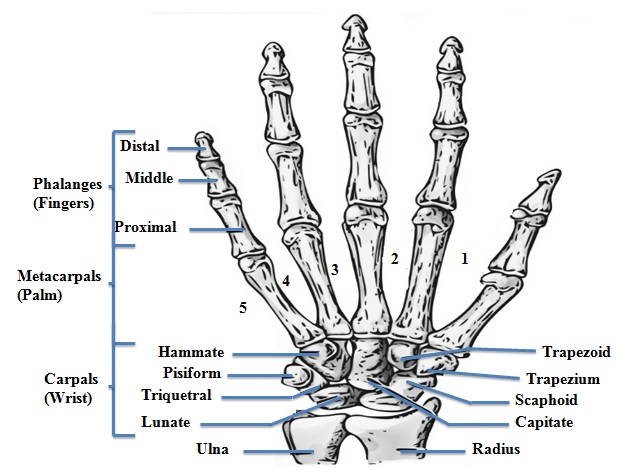}
\caption{Bones in Human hand, their names and classification.}
\label{fig:hand}
\end{figure}
 
 \citep{afifi17} used Convolutional Neural Networks to determine gender of the individuals by biometric tracts found in hands. They used the images of hands on both sides (Palm and Dorsal). They claim to achieve a good accuracy not only on palm side but also on dorsal side of the hand.
\citep{mfdar15} used a Hybrid Particle Swarm Artificial Neural Network technique to determine the gender of individuals which is most relevant work to our research. They used a dataset of left hand X-ray images of Asian population. Their results suggest a different accuracy for different age group. Their dataset was small and they never mentioned the area of attention in their research work. 

\citep{scdpin16} introduced a method for objective quantification of sexually dimorphic features using wavelet transform on the images of skull and pelvis. They claimed that their method had been successfully applied for gender determination of pilot sample of 3D meshes which are a kind of digital record of supraorbital morphology. 

\citep{habdul17} used micro structural image processing to identify gender from a sample of bones. They divided their detection system into two parts. In the first part they manually analyzed and observed the differences in the parameters of male and female samples of bones. In the second part of their system they applied the image processing techniques to identify the gender.

\citep{garvin12} presented a technique for quantification of morphological variation using three-dimensional surface lasers in the skeletal brow ridge and chin. They isolated the chin and brow ridge regions on objectively defined planes and landmarks. Authors used volumes and region areas to quantify relative and absolute sizes. The authors also considered the semi landmarks for further morph metric analysis. They claimed by their results that their technique can quantify chin and  brow ridge morphologies which can easily be analyzed using statistical methods.

The techniques used previously mainly focus the gender determination of adults. The body of an adult is mature enough to point out significant clues which can help to distinguish the gender related features in a male or a female. But the approaches of gender determination for adults may not be suitable for gender determination of children, therefore this research is focused on gender determination of the children using only the radio-graphs of their left hand. 
 \begin{figure*}[h]
\includegraphics[width=\linewidth]{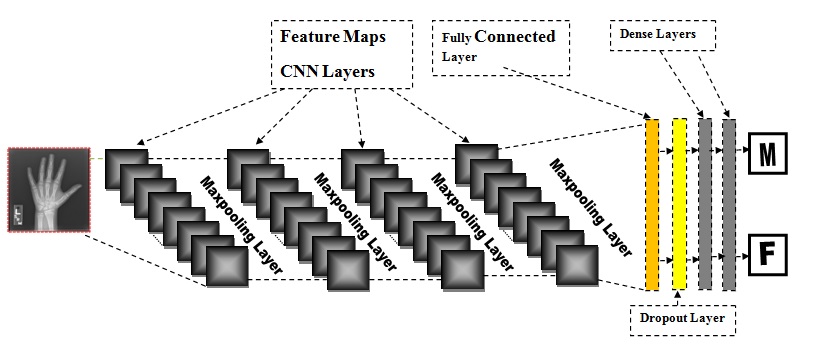}
\caption{The architecture/layout of the proposed CNN model.}
\label{fig:arccnn}
\end{figure*}
\begin{figure*}[]
\includegraphics[width=\linewidth]{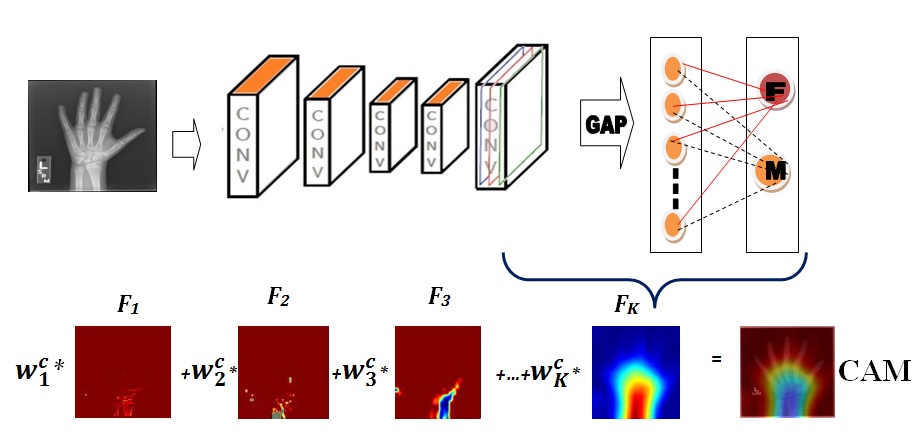}
\caption{Class Activation Mapping of the proposed model.}
\label{fig:cam}
\end{figure*}

\section{Preliminary}
\subsection{Problem and Motivation}
In ideal situation the techniques used by the previous researchers may be suitable to be used for gender determination. But these techniques may fail in scenarios where there is only a small portion of the body (without or with damaged skull and pelvis area). If there is a huge chance of physical, biological or  a chemical damage to the \textit{remains} of the subject the previous techniques may also fail. Specially the chemical reactions to body are mostly irreversible and they can make a forensic investigation impossible to determine the gender. In such a situation bones in the hand of a subject as shown in Figure~\ref{fig:hand} may play a crucial role in forensic investigation to determine the gender. With age the bones in the human body grow proportionally to each other and growth is also gender specific.Therefore bones in the hand of a child have different stature as compared to an adult. In case the subject is a child it is very difficult to differentiate between the hand of a male and female child only by looking with naked eye. The manual measurements or physical examinations of the hand (which may be burned due to a disaster) is very complex. In most of the cases the physical examination is not possible or may produce inaccurate results.\\

In recent years deep learning has become a standard in almost all the fields of computer vision and related problems \citep{bwang17}, \citep{david17}. In challenging fields such as bioinformatics and medical imaging the deep learning models has a huge role \citep{kskur17}, \citep{bzhu16}, \citep{dbard18}. Deep learning not only requires a huge amount of training data but also a reasonable memory size and sufficient computational resources to acquire acceptable accuracy, it was very difficult in past to meet such a challenge \citep{mhu11}, \citep{sakcay18}. But recently this challenge has been neutralized thanks to high performance GPUs and big data generating digital platforms \citep{fskhan14}, \citep{inyu17}. A number of architectures and models have been devised to achieve accuracy of almost one hundred percent \citep{jzhou17}. ConvNets or CNNs (Convolutional Neural Networks) as depicted in Figure~\ref{fig:arccnn} have proved to be one of the main concepts in deep learning. Convolutional neural networks usually consist of many layers \citep{david17}. The layers can be convolutional layers, maxpooling layers and fully or densely connected layers. Broadly speaking there are two major concepts called as forward propagation and backward propagation. In both the concepts the working principle for the inner layers will be different. In this research we propose to use CNNs in order to determine the gender of the subjects. The objective of the proposed technique is to classify the X-ray images of the left hands of children in the age of one month to 18 years old.\\
As visualizing the network during and after training is very important in order to recognize the patterns the network is learning or have learned. One of the techniques is the procedure of generating class activation maps (CAM) \citep{bzhu16} using global average pooling (GAP) \citep{bzhu16} in convolutional neural networks. A CAM indicates the descriminative regions of an image for a dedicated class used by a CNN in order to identify the class. Figure~\ref{fig:cam} illustrates the procedure of the CAM. 

The network based on convolutional layers is very much suitable for class activation mapping .In such a network exactly before the softmax layer if it is a classification problem, a method called as global average pooling (GAP)\citep{bzhu16} is performed. The convolutional feature maps are used by the GAP and then the desired output is produced using such features in fully-connected layer. With this straightforward connectivity structure, the weights are projected back  on the convolutional feature maps to identify the important image regions. It can be seen in Figure~\ref{fig:cam}, that the GAP is producing the spatial average which is calculated over the feature maps on each unit of the final convolutional layer.

To generate the final output by using the weighted sum of the computed values. A weighted sum of the feature maps of the final convolutional layer is computed to acquire a class activation map (CAM). For an input image X, we consider the activations of the last convolutional layer in the network. For each featuremap $k$, let $f_k(x, y)$ be the activation value at spatial point $(x,y)$, we perform GAP to get the summation of all the values on featuremap $k$,$F_k =\Sigma_{x,y} f_k (x,y)$, as the output of the network.\\
Then we connect the output of GAP to the final class layer. Let $w_k^c$ be the weight connecting unit $k$ to class $c$. Then $S_c = \Sigma_k w_k^c F_k$ is the input to the softmax for class $c$ as shown in Figure~\ref{fig:cam}, and finally $P_c = \frac{exp(S_c)}{\Sigma_c exp(S_c)}$ is the output  of the softmax layer for class $c$. Bias term is ignored as it has no impact.
 \begin{equation}
S_c= \Sigma_k w_k^c \Sigma_x,_y f_k(x,y) =\Sigma_x,_y \Sigma_k w_k^c f_k (x,y)\label{eq:01}\vspace*{-10pt}
\end{equation}
The class activation map for class $c$ is defined as $M_c$, and  each spatial point is given by
 \begin{equation}
M_c(x,y)= \Sigma_k w_k^c f_k(x,y)\label{eq:02}\vspace*{-10pt}
\end{equation}

Therefore, $S_c = \Sigma_x,_y M_c(x,y)$ and $M_c(x, y)$ directly points out the significance of the activation map at spatial point $(x, y)$ which leads to classify the input image for the class $c$. Each unit is activated because of some pattern within its attentive region. Hence $f_k$ presents the map of the pattern. The CAM is basically a weighted linear sum of visual patterns at different spatial locations. Relevant image regions can be identified on input image by simply up sampling the class activation maps.
By up sampling the class activation map to the size of the input image, image regions most relevant to the particular class can be identified. 

\subsection{Anatomy of Human Hand}
Literature suggests that a human hand have significant clues to identify the gender of the subject comprehensively \citep{david17}. Therefore this research is focused on the gender determination based on X-ray images of the hands of children. From Figure~\ref{fig:hand} it can be seen that a human hand has different kinds of bones at different levels. There are three major portions mainly Phalanges (Fingers), Metacarpals (Palm) and Carpals (Wrist). The two bones connecting the arm with hand  are called Ulna and Radius. The phalanges area has three bones named as Proximal, Middle and Distal respectively. The Metacarpals area has bones which are not only stronger but they also serve as a foundation and bridge between Phalanges and Carpals \citep{edwards13}. The carpals area contains eight bones Hammate, Pisiform, Triquetral, Lunate, Trapezoid, Trapezium, Scaphoid, and Capitate. These bones make a grid around wrist area to facilitate the free movement of the hand. This anatomy of human hand may be different in a male child as compared to a female child.

\section{Gender Determination with CNN}
\begin{figure}[h]
\includegraphics[width=\linewidth]{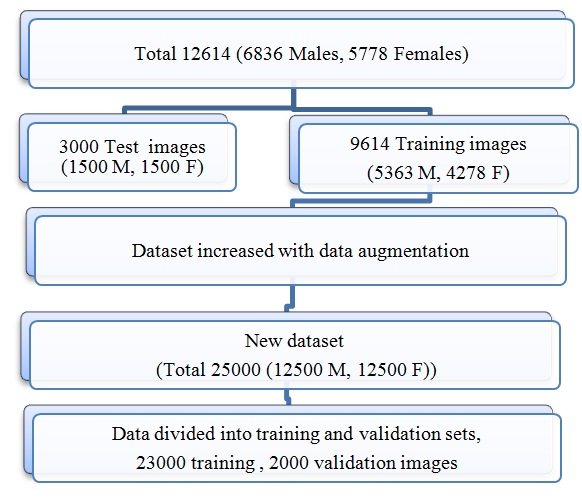}
\caption{Training and testing set creation and distribution.}
\label{fig:dataset}
\end{figure}

The main component of every deep learning model is usually the dataset.The dataset has been taken from the web resource provided by the Stanford Center for Artificial Intelligence in Medicine \& Imaging \citep{david17}. The major contributors in this dataset are Stanford University, the University of California - Los Angeles and the University of Colorado. This dataset was used in the Pediatric Bone Age Challenge Competition \citep{david17}. The details about the dataset are given in Figure~\ref{fig:dataset}. As most of the deep learning models require huge number of training images to produce desirable results, we applied data augmentation technique by adding Gaussian noise \citep{rmzur09} to augment the dataset. The increased dataset has a total of 25000 images as depicted in Figure~\ref{fig:dataset}. The dataset consists of high resolution gray scale images.

\begin{figure*}[h]
\includegraphics[width=\linewidth]{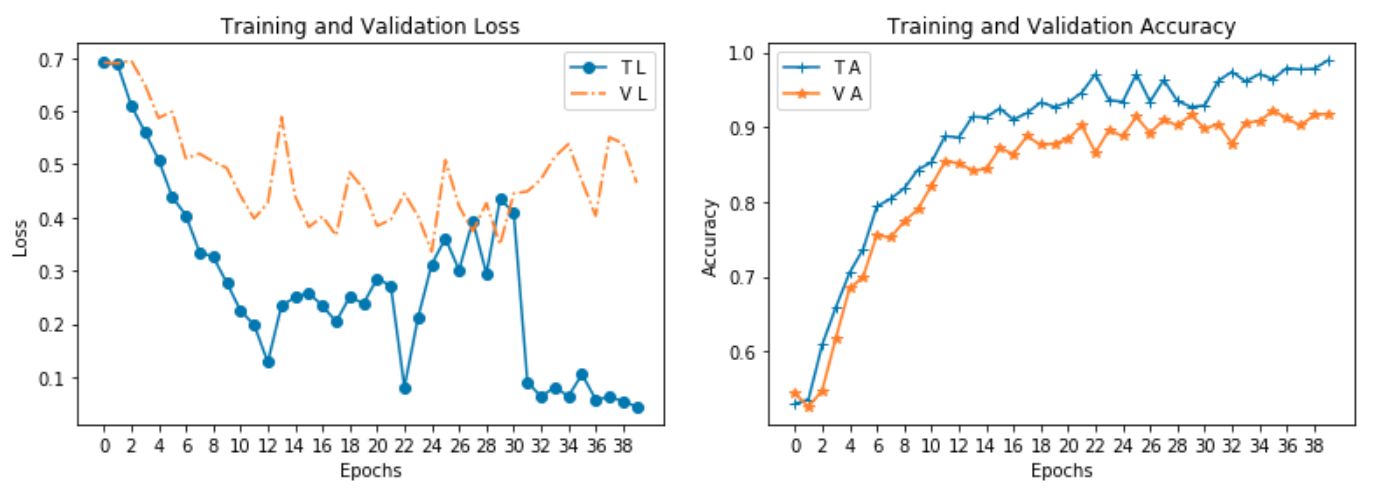}
\caption{Graphs showing training accuracy and validation accuracy of the model. The training loss and validation loss are also depicted (Here TL= Training Loss, VL= Validation Loss, TA=Training Accuracy, VA=Validation Accuracy).}
\label{fig:acculoss}
\end{figure*}
Before training the network the data has been preprocessed using openCV with python in order to make the dataset big enough that can help to achieve optimal results. The testing images has been selected randomly for both the genders. As the ages of the subjects are between one month and 18 years, therefore dataset covers almost all the age groups for both the genders. After selecting the test dataset the rest of the images have been preprocessed by adding Gaussian noise \citep{rmzur09}. The reason for preprocessing was to avoid over-fitting of the model.

\subsection{Network Architecture}

Architecture of the proposed GDCNN model is based on deep convolutional neural networks, the block diagram of the model is depicted in Figure~\ref{fig:arccnn}. Following is the summary of the model.
\begin{itemize}
\item There are four convloutional layers, the size of the input image is (137, 137), kernel size is (3, 3), activation function is Relu, a filter of (2, 2) filter has been used for maxpooling after every convolutional layer.
\item There are two dense layers, the first dense layer  uses Relu as activation function and last dense layer uses Sigmoid as activation function
\item Dropout ratio of 0.8 has been used to avoid over-fitting
\item Adams algorithm has been used as an optimizer and binary cross entropy has been used as a loss function.
\vspace*{1pt}
\end{itemize}
\subsection{Training and Validating the Network}
The network has been implemented using TensorFlow 1.7 on a Ubuntu 16.04 powered with 3 GPU (Geforce GTX TitanX) CUDA 8 cudNN 5.1, having RAM of 62.8 GB. All the data pre-processing tasks were performed using Python 3.6.
The proposed model consists of two major parts as illustrated in Figure~\ref{fig:arccnn} and Figure~\ref{fig:cam}. The first part has the goal of training and validation, and the second part uses class activation mapping (CAM) to identify the area of attention.
In this research we use TensorFlow 1.7 to train the network, therefore it is very important to reshape the data into a tensor so that it may be used to train the network. Data is usually fed in batches to train the network in order to avoid any system crashes or decreased performance.While training we used comparatively larger image size so the batch size is set to only 50 images and step size is 5000.The number of epochs to train the network has been set to 40. As it is evident from Figure~\ref{fig:acculoss} the model has a reasonable training accuracy and validation accuracy which can be observed from the graphs of accuracy and loss. After the final epoch the training accuracy has been noted as 0.979, validation accuracy as 0.918, the training loss as 0.055 and the validation loss as 0.465. The accuracy of the model is very important to make correct predictions later on the test dataset. 
\begin{methods}
\section{Experiments and Analysis}
We implemented the model to test in a real world scenario in order to check its feasibility to deploy. The test images were already selected before any prepossessing, as depicted in Figure~\ref{fig:dataset}. The test images have never been used in training and validation process.

\subsection{Evaluation Metrics}
To evaluate the model with standard procedure the true positive (TP), true negative (TN), false positive (FN) and false negative (FN) values were determined. On the basis of these values accuracy, precision, recall and F1-score \citep{kskur17} were calculated.

Accuracy is the most intuitive performance measure and it is simply a ratio of correctly predicted observation to the total observations. The accuracy of the model on the basis of $TP, TN, FP$ and $FN$ is calculated using ratio as given in Equation 3.
\begin{equation}
accuracy = \frac{TP+TN}{TP+FP+TN+FN}\label{eq:03}\vspace*{-10pt}
\end{equation}
Recall is the ratio of correctly predicted positive observations to the all observations in actual class - yes. The ratio is given in Equation 5.
\begin{equation}
recall= \frac{TP}{TP+FN}\label{eq:05}\vspace*{-10pt}
\end{equation}
Precision is the ratio of correctly predicted positive observations to the total predicted positive observations as given in Equation 4.
\begin{equation}
precision = \frac{TP}{TP+FP}\label{eq:04}\vspace*{-10pt}
\end{equation}
 F$_1$-Score is the weighted average of Precision and Recall. The ratio is given in Equation 6.
\begin{equation}
F_1= \frac{precision}{recall+precision}\label{eq:06}\vspace*{-10pt}
\end{equation}

 \subsection{Computing and Comparing the Results}

There are 3000 images in the test dataset which have equal number of male and female samples (1500 for each class). The trained model was tested for predictions on the test dataset. As depicted in Table~\ref{Tab:01} the model has a high number of True Positives and True Negatives for both classes. It indicates the model has generalized well to make correct predictions. On the basis of $TP, FP, FN$ and $TN$, the accuracy, precision, recall and $F_1$ scores have been calculated. The results show a satisfactory performance as compared to previous techniques.~Table~\ref{Tab:02} shows the most recent work for gender determination. It can be seen that the proposed model performs better than state-of-the-art techniques.

\begin{table}[!t]
\processtable{Results obtained after making predictions on the trained model, \label{Tab:01}} {\begin{tabular}{@{}lllllllll@{}}\toprule
Class & TP & FP & FN & TN & A & P & R & F$_1$\\\midrule
Male & 1455 & 30 & 45 & 1470 & 0.975 & 0.979 & 0.970 & 0.974\\
Female & 1467 & 47 & 33 & 1453 & 0.973 & 0.968 & 0.978 & 0.972\\\botrule
\end{tabular}}{A=Accuracy, P=Precision, R=Recall, F$_1$=F$_1$ Score}
\end{table}
\begin{table}[!t]
\processtable{Accuracy of the previous techniques compared with proposed technique\label{Tab:02}} {\begin{tabular}{@{}lll@{}}\toprule
Author & Male & Female\\\midrule
\citep{ibrahim16} & 0.960 & 0.940 \\
\citep{aboulh11} & 0.888 & 0.804 \\
\citep{thomas18} & 0.680 & 0.680 \\
\citep{fgiu13} & 0.890 & 0.870 \\
\citep{edwards13} & 0.660 & 0.680 \\
\citep{afifi17}$^P$ & 0.912 & 0.835 \\
\citep{afifi17}$^D$ & 0.919 & 0.898 \\
\citep{mfdar15} & 0.722 & 0.694 \\
\citep{garvin12}$^B$ & 0.818$^*$ & 0.967$^*$ \\
\citep{garvin12}$^B$ & 0.967$^+$ & 0.923$^+$ \\
\citep{garvin12}$^C$ & 0.818$^*$ & 0.667$^*$ \\
\citep{garvin12}$^C$ & 0.767$^+$ & 0.731$^+$ \\
\bf{Proposed Model} & \bf{0.975} & \bf{0.973} \\\botrule
\end{tabular}}{The accuracy of the previous techniques lower as compared to the proposed model, P=Palm Side of hand has been used, D=Dorsal Side of hand has been used, B=Brow Ridge, C= Chin, *= Samples of Black People, += Sample of White People}
\end{table}

\begin{figure*}[]
\includegraphics[width=\linewidth]{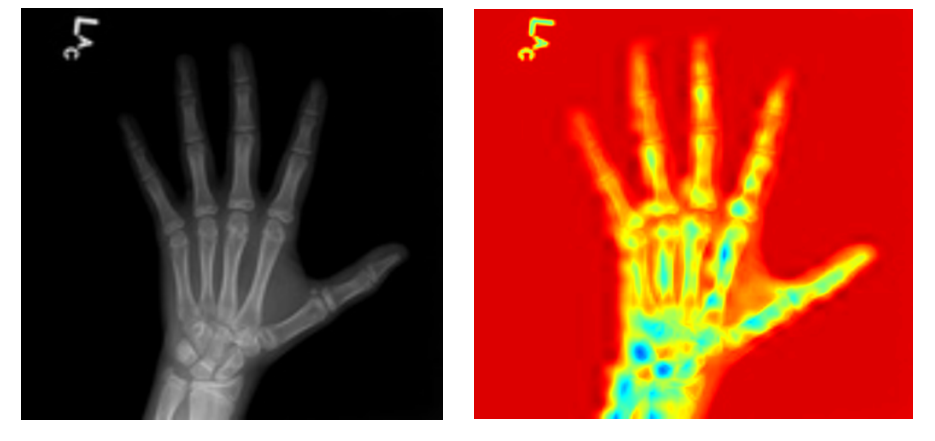}
\caption{The original input image correctly predicted as a female child's hand and resultant output image is clearly depicting the area under attention.}
\label{fig:femres}
\end{figure*}
\begin{figure*}[]
\includegraphics[width=\linewidth]{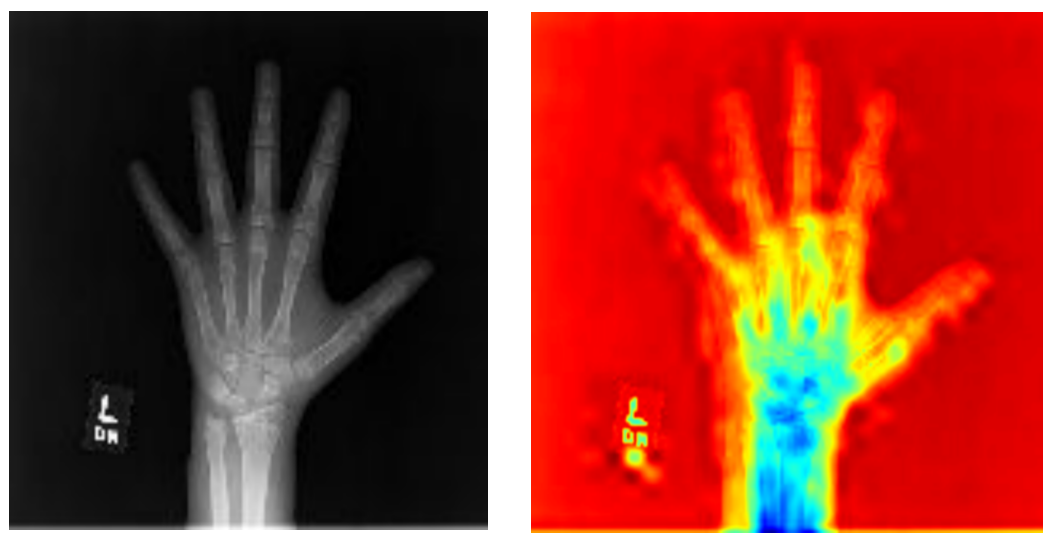}
\caption{The original input image correctly predicted as a male child's hand and resultant output image is clearly depicting the area under attention.}
\label{fig:malres}
\end{figure*}
\subsection{Further Exploration of Area of Attention}
To identify the area of attention the class activation mapping (CAM) has been used, which highlights the area where the feature maps are focused. The area of attention for all the test images has been determined by observing heatmaps generated for each of the input test image. As depicted in Figure~\ref{fig:femres} and Figure~\ref{fig:malres}, most of the attention of the model in both cases (males and females) is around Carpals (wrist) area, which is a junction between bones of Ulna, Radius, and Metacarpals.
As Carpals is a small gird of bones (Hammate, Pisiform, Triquetral, Lunate, Trapezoid, Trapezium, Scaphoid and Capitate), the network has learned much of the gender based features from this area.

Figure~\ref{fig:attention} shows the network is more focused around lower part of the hand. For almost all the test images the focus is around the Carpals. Figure~\ref{fig:attention} depicts that more than 1400 test images for both male and females share an attention around Carpals and Radius, more than 1200 test images has attention around Ulna, Carpals and Radius, more than 1000 images share an attention around Ulna, Carpals, Radius and Metacarpals. Only about 400 images share attention around all the regions of the hand.
This is a clear indication that the learned model is more focused around the Carpals region.
The results indicate that Carpals region and Radius bones are more important to determine the gender of the children. In younger age the bones in the upper parts of the hands may be identical in both genders, therefore more focus is around lower part of the hand. Another important pattern has been noted that the bones in Carpals region of the males are thicker and are very close to each other while in females the bones look thinner and there are visible gaps between the bones around this area. This can be clearly observed from the heatmap generated with CAM as shown in Figure~\ref{fig:femres} and Figure~\ref{fig:malres}. The fingers (Phalanges) in the hands of children play a very little role in gender determination.\vspace*{1pt}
Table~\ref{Tab:02} shows that the proposed model has a higher accuracy to predict the gender even if the subject is a child. Though most of the previous techniques were applied to determine the gender of adults, they still have a lower accuracy as compared to the proposed model.

\begin{figure*}[]
\includegraphics[width=\linewidth]{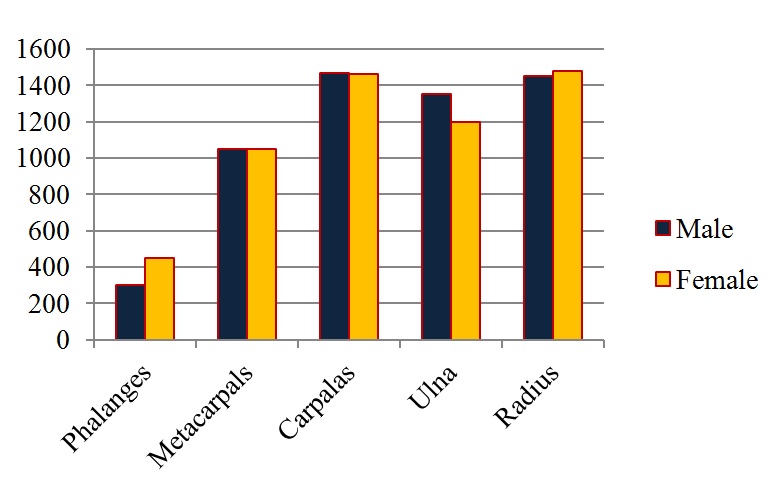}
\caption{Area of attention for test dataset. It can be clearly seen that most of the attention is around carpals, ulna and radius for both the genders.}
\label{fig:attention}
\end{figure*}
\end{methods}

\section{Conclusion}
We proposed a deep learning based model which can determine the gender of a child simply with a hand radio-graph. The proposed model can also identify the area of attention which can further confirm the correct prediction. The previous techniques were lacking not only in terms of accuracy but also in terms of authenticity. The proposed model can identify the gender of a child even with a half portion of lower part of the hand (Carpals). But the previous techniques may not perform well with a half sample. 
%
%

The deep learning techniques are very much suitable for digital image forensics, especially when there is a reasonable dataset available to train, validate and test the network. In this paper we successfully trained and tested the network with a high accuracy on child gender determination only on left hand radio-graphs.
The network predicted the gender with an acceptable accuracy. It was also observed with class activation mapping (CAM) that the bones at carpals, ulna and radius are more important to determine the gender of the children using left hand radio-graphs. We used left hand radio-graphs as only such dataset is available on the website. In future the model may also be trained using not only left hand images of the children but it can also be trained with the X-ray images of right hands, to find a definite connection or differences in the bone structure in the hands of children, which may further help the forensic investigators in gender determination. Further the hand radio-graphs of adults can also be used to train a network to test its efficacy in gender determination.\vspace*{-10pt}

\section{Funding}

This work is supported by Chinese National Natural Science Foundation (61772219).\vspace*{-12pt}

%
%

\begin{thebibliography}{}
\bibitem[Thomas Radulesco {\it et~al}., 2018]{thomas18}
Thomas Radulesco et al. (2018) Sex Estimation from Human Cranium: Forensic and Anthropological Interest of Maxillary Sinus Volumes. {\it Journal of Forensic Sciences} {\bf 63}, no. 3, 805-808.


\bibitem[David B. Larson {\it et~al}., 2017]{david17}
David B. Larson {\it et~al} (2017) Performance of Deep- Learning  Neural   Model in Assessing Skeletal Maturity on Pediatric Hand Radiographs . {\it Radiology Vol. 287}.

\bibitem[Afifi {\it et~al}., 2017]{afifi17}
Afifi, Mahmoud. {\it et~al} (2017), Gender Recognition and Biometric Identification Using a Large Dataset of Hand Images. arXiv preprint arXiv:1711.04322.

\bibitem[H. Abdullah {\it et~al}., 2017]{habdul17}
H. Abdullah {\it et~al} (2017)  Automated Haversian Canal Detection for Histological Sex Determination, {\it IEEE Symposium on Computer Applications \& Industrial Electronics (ISCAIE), Langkawi}, 69-74.

\bibitem[M. F. Darmawan  {\it et~al}., 2015]{mfdar15}
 M. F. Darmawan {\it et~al}. (2015) Hybrid PSOANN for sex estimation based on length of left hand bone, {\it IEEE Student Conference on Research and Development (SCOReD), Kuala Lumpur}, 478-483.
 
 \bibitem[S. C. D. Pinto {\it et~al}., 2016]{scdpin16}
 S. C. D. Pinto {\it et~al}. (2016) Two-Dimensional Wavelet Analysis of Supraorbital Margins of the Human Skull for Characterizing Sexual Dimorphism, {\it IEEE Transactions on Information Forensics and Security, vol. 11}, {\bf 7},1542-1548.
 
 \bibitem[Ibrahim {\it et~al}., 2016]{ibrahim16}
Ibrahim {\it et~al}. (2016) Sex Determination From Hand Dimensions and Index/Ring Finger Length Ratio in North Saudi population: Medico-legal View, {\it Egyptian Journal of Forensic Sciences},{\bf 4}, 435-444.

\bibitem[F. Giurazza {\it et~al}., 2013]{fgiu13}
 F. Giurazza {\it et~al}. (2013) Sex determination from scapular length measurements by CT scans images in a Caucasian population, {\it 35th Annual International Conference of the IEEE Engineering in Medicine and Biology Society (EMBC)}, Osaka, 1632-1635.
 
 \bibitem[Aboul Hagag {\it et~al}., 2011]{aboulh11}
Aboul-Hagag  {\it et~al}. (2011) Determination of gender from hand dimensions and index/ring finger length ratio in Upper Egyptians.{\it Egyptian Journal of Forensic Sciences},{\bf 1}, 80-86.

\bibitem[Garvin {\it et~al}., 2012]{garvin12}
Garvin {\it et~al}. (2012) Sexual dimorphism in Skeletal Brow Ridge and Chin morphologies determined using a new quantitative method, {\it American Journal of Physical Anthropology},{\bf 147}, no. 4, 661-670.

\bibitem[Edwards {\it et~al}., 2013]{edwards13}
Edwards {\it et~al}. (2013) Sex Determination from the Foramen Magnum. {\it Journal of forensic radiology and imaging}, {\bf 1}, no. 4, 186-192.

\bibitem[J. Zhou {\it et~al}., 2017]{jzhou17}
J. Zhou  {\it et~al}. (2017) Using Convolutional Neural Networks and Transfer Learning for Bone Age Classification, {\it International Conference on Digital Image Computing: Techniques and Applications (DICTA)}, Sydney, NSW,  1-6.

\bibitem[F. S. Khan {\it et~al}., 2014]{fskhan14}
 F. S. Khan {\it et~al}. (2014) Semantic Pyramids for Gender and Action Recognition, {\it IEEE Transactions on Image Processing}, {\bf 23}, no. 8, 3633-3645.
 
\bibitem[M. Hu {\it et~al}., 2011]{mhu11}
M. Hu {\it et~al}. (2011) Gait-Based Gender Classification Using Mixed Conditional Random Field,  {\it IEEE Transactions on Systems, Man, and Cybernetics, Part B (Cybernetics)},{\bf 41}, no. 5,1429-1439.

\bibitem[I. N. Yulita {\it et~al}., 2017]{inyu17}
I. N. Yulita {\it et~al}. (2017) Sleep Stage Classification Using Convolutional Neural Networks and Bidirectional Long Short-Term Memory, {\it International Conference on Advanced Computer Science and Information Systems (ICACSIS)}, Bali, 303-308.

\bibitem[D. Bardou {\it et~al}., 2018]{dbard18}
D. Bardou {\it et~al}. (2018) Classification of Breast Cancer Based on Histology Images Using Convolutional Neural Networks, {\it IEEE Access}, {\bf 6}, 24680-24693.

\bibitem[S. Akcay {\it et~al}., 2018]{sakcay18}
S. Akcay {\it et~al}. (2018) Using Deep Convolutional Neural Network Architectures for Object Classification and Detection Within X-Ray Baggage Security Imagery, {\it IEEE Transactions on Information Forensics and Security},{\bf 13}, no.9, 2203-2215.

\bibitem[K. S. Kurachka {\it et~al}., 2017]{kskur17}
K. S. Kurachka {\it et~al}.  (2017) Vertebrae Detection in X-ray Images Based on Deep Convolutional Neural Networks, {\it IEEE 14th International Scientific Conference on Informatics, Poprad}, 194-196.

\bibitem[B. Wang {\it et~al}., 2017]{bwang17}
B. Wang {\it et~al}. (2017) X-Ray Scattering Image Classification Using Deep Learning, {\it IEEE Winter Conference on Applications of Computer Vision (WACV)}, Santa Rosa, CA,  697-704.


\bibitem [B. Zhu {\it et~al}.,(2016)]{bzhu16}
B. Zhu {\it et~al}. (2016) Learning Deep Features for Discriminative Localization, {\it IEEE Conference on Computer Vision and Pattern Recognition (CVPR)}, Las Vegas, NV, 2921-2929.

\bibitem [R.M. Zur {\it et~al}.(2009)]{rmzur09}
R.M. Zur {\it et~al}. (2009), Noise Injection for Training Artificial Neural Networks: A comparison with weight decay and early stopping, {\it Med. Phys, 36(10)}.


\end{thebibliography}

\end{document}